\documentclass[letterpaper]{article} 
\usepackage{aaai23}  
\usepackage{times}  
\usepackage{helvet}  
\usepackage{courier}  
\usepackage[hyphens]{url}  
\usepackage{graphicx} 
\usepackage{natbib}  
\usepackage{caption} 
\usepackage{algorithm}

\usepackage{newfloat}
\usepackage{listings}
\DeclareCaptionStyle{ruled}{labelfont=normalfont,labelsep=colon,strut=off} 
\floatstyle{ruled}
\newfloat{listing}{tb}{lst}{}
\floatname{listing}{Listing}
\usepackage{amsmath,amsfonts,bm}

\def\eqref#1{equation~\ref{#1}}

\def\1{\bm{1}}

\def\vtheta{{\bm{\theta}}}

\def\vf{{\bm{f}}}
\def\vg{{\bm{g}}}

\def\vx{{\bm{x}}}
\def\vy{{\bm{y}}}

\def\mF{{\bm{F}}}

\def\mX{{\bm{X}}}

\DeclareMathAlphabet{\mathsfit}{\encodingdefault}{\sfdefault}{m}{sl}
\SetMathAlphabet{\mathsfit}{bold}{\encodingdefault}{\sfdefault}{bx}{n}

\DeclareMathOperator*{\argmin}{arg\,min}

\usepackage{algpseudocode}
\usepackage{subcaption}
\usepackage{array}

\newcommand{\PreserveBackslash}[1]{\let\temp=\\#1\let\\=\temp}
\newcolumntype{C}[1]{>{\PreserveBackslash\centering}p{#1}}
\newcolumntype{R}[1]{>{\PreserveBackslash\raggedleft}p{#1}}
\newcolumntype{L}[1]{>{\PreserveBackslash\raggedright}p{#1}}

\title{Boosted Dynamic Neural Networks}
\author{
    Haichao Yu$^{1}$, Haoxiang Li$^{2}$, Gang Hua$^{2}$, Gao Huang$^{3}$, Humphrey Shi$^{1,4}$\\
}
\affiliations{
    $^{1}$UIUC, $^{2}$Wormpex AI Research, $^{3}$Tsinghua University, $^{4}$University of Oregon
}

\usepackage{bibentry}

\begin{document}

\maketitle

\begin{abstract}
Early-exiting dynamic neural networks (EDNN), as one type of dynamic neural networks, has been widely studied recently. A typical EDNN has multiple prediction heads at different layers of the network backbone. During inference, the model will exit at either the last prediction head or an intermediate prediction head where the prediction confidence is higher than a predefined threshold. To optimize the model, these prediction heads together with the network backbone are trained on every batch of training data. This brings a \textit{train-test mismatch} problem that all the prediction heads are optimized on all types of data in training phase while the deeper heads will only see difficult inputs in testing phase. Treating training and testing inputs differently at the two phases will cause the mismatch between training and testing data distributions. To mitigate this problem, we formulate an EDNN as an additive model inspired by gradient boosting, and propose multiple training techniques to optimize the model effectively. We name our method BoostNet. Our experiments show it achieves the state-of-the-art performance on CIFAR100 and ImageNet datasets in both anytime and budgeted-batch prediction modes. Our code is released at https://github.com/SHI-Labs/Boosted-Dynamic-Networks.
\end{abstract}
\section{Introduction}

Despite the success of the deep learning models on various computer vision tasks, computational resources limitation and inference efficiency must be considered when deploying the models in real-world applications. As a crucial topic in deep learning application, efficient deep learning has been extensively investigated recently, including efficient architecture design~\cite{iandola2016squeezenet,howard2017mobilenets,sandler2018mobilenetv2,zhang2018shufflenet,han2020ghostnet}, network pruning~\cite{NIPS2015_ae0eb3ee,han2015deep_compression,li2016pruning,he2017channel,luo2017thinet,he2019filter,lin2020hrank}, and network quantization~\cite{courbariaux2016binarized,rastegari2016xnor,zhou2016dorefa,li2016ternary,zhang2018lq,liu2018bi,wang2019haq,yu2019any,liu2021post}. Compared with the corresponding large baseline models, these lightweight models usually exhibit significant inference speed-up with small or negligible performance drop.

Dynamic neural networks is another branch of efficient deep learning models. Practically, lightweight models run faster with lower performance, while large models run slower with higher performance. Given a computational resource budget in a specific application scenario, deep learning practitioners often need to find a proper trade-off point between efficiency and performance. However, the budget is often dynamically changed in many scenarios, for example, when the computational resources are largely occupied by other applications or an edge-device's battery level changes in a day. Therefore, a dynamic neural network (DNN) is desirable that can run at different trade-off points given different resource budgets and performance requirements. Extensive research has been conducted on various dynamic methods including dynamic model architectures, dynamic model parameters, dynamic spatial-wise inference, and more~\cite{teerapittayanon2016branchynet,wang2018skipnet,veit2018convolutional,wu2018blockdrop,yu2019any,li2017dynamic,wang2019elastic,wang2020glance,yang2020resolution,han2021dynamic}. In this paper, we will focus on early-exiting dynamic neural network (EDNN) on image classification tasks, one of most widely explored dynamic architecture methods~\cite{huang2017multi,mcgill2017deciding,li2019improved,jie2019anytime,yang2020resolution}.

EDNN's adaptive inference property comes from its adjustable layer depth. In inference, when the computational budget is tight or the input example is ``easy'' to recognize, EDNN exits at a shallow classifier. In the opposite, the model runs through more layers until the last classifier or the classifier that gives a confident prediction. To train such an EDNN model, existing methods usually employ joint optimization over all the classifiers~\cite{huang2017multi,li2019improved,yang2020resolution}. Given a batch of training samples, the model runs through all the layers and classifiers, and the losses are accumulated and summed up. Then the gradients of the total loss are back-propagated to all the layers, and the model parameters are updated accordingly.

One issue in the above training procedure is the \textit{train-test mismatch} problem, which is also discussed in \cite{han2021dynamic}. In the training stage, all the classifiers in the model are optimized over all the training samples, while in the inference stage, not all the classifiers will see all types of testing data: when the budget is tight or the input is ``easy'' to process, only shallow layers and classifiers will be executed. In other words, there exists a data distribution gap between the training and testing stages. Although the ``easy'' examples may also help regularize the deep classifiers in the training stage, the deep classifiers should not be designed to focus too much on these examples in order to avoid the distribution mismatch problem.

To mitigate the {train-test mismatch} problem, we propose a new type of early-exiting dynamic networks named Boosted Dynamic Neural Networks (BoostNet), inspired by the well-known gradient boosting theory. Assuming the final model is a weighted sum of weak prediction models, gradient boosting algorithms minimize the loss function by iteratively training each weak model with gradient descent in function space. In each training step, the model that points to the negative gradient of the loss function is selected as a new weak model. In this way, each weak model is supplementary to the ensemble of its previous weak models. This property is what we desire in training an EDNN model.

\begin{figure}[t]
    \centering
    \includegraphics[width=0.98\linewidth]{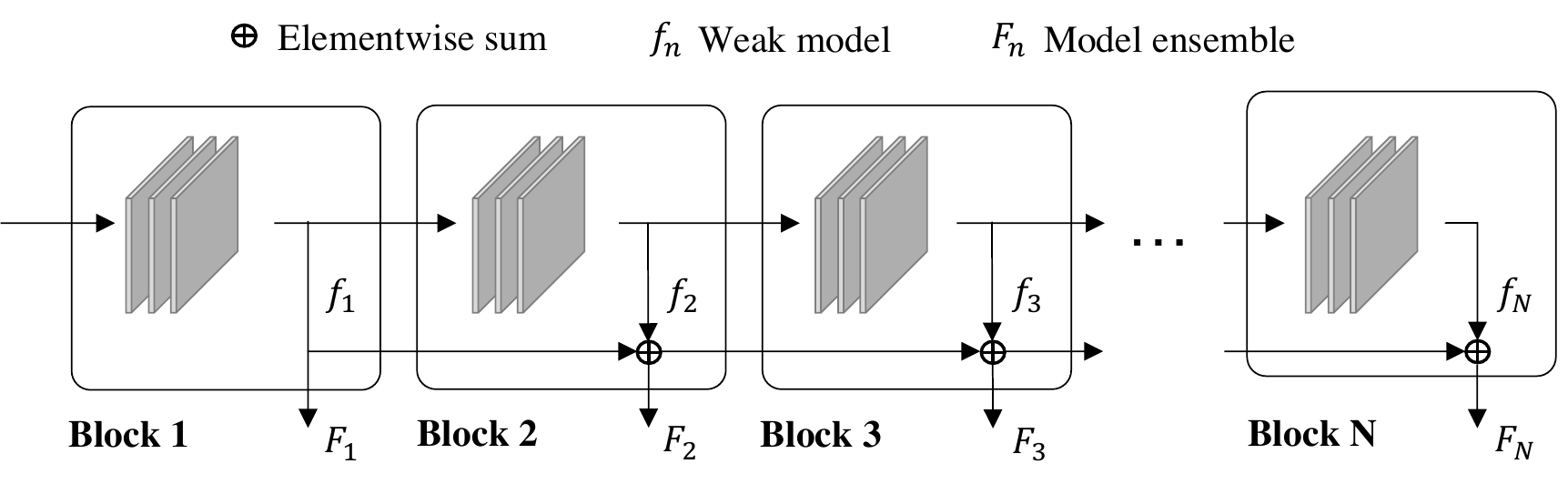}
    \caption{The architecture of the proposed Boosted Dynamic Neural Networks (BoostNet).}
    \label{fig:arch}
\end{figure}

A straightforward manner to incorporate gradient boosting into EDNN training is to directly follow the conventional gradient boosting pipeline to iteratively train the classifiers in the network. However, this does not work well in practice. We speculate there are two main reasons. The first reason is that the deeper classifiers share network layers with the shallower classifiers, so the classifiers are not independent in terms of the network parameters\footnote{One may argue to disentangle the classifiers in the EDNN into independent ones. Compared with backbone sharing, training independent classifiers is not efficient in the aspects of model size and computational complexity.}. The second reason is that a multi-layer neural network is not a weak prediction model. These two properties do not comply with the assumptions of gradient boosting. Therefore, na\"ively training the classifiers in a gradient boosting way will not work. In this paper, we propose BoostNet, which shares the similar spirits of gradient boosting but is organized and trained in a manner adapted to dynamic neural networks. Refer to Fig.~\ref{fig:arch} for an architecture overview of the proposed BoostNet, where the classifiers are organized as a dynamic-depth neural network with early exits.

To train the model effectively, we propose three training techniques. First, we employ joint optimization over all the classifiers. Training the weak classifiers in the gradient boosting way will lead to a sub-optimal solution for the whole model. On the other hand, training the classifiers sequentially on each training batch is time-costly. Therefore, we train all the classifiers via joint optimization.
Second, we propose a prediction reweighting strategy to rescale the ensemble of the previous classifiers before being used an ensemble member of the current classifier. We define a training sample as a \textit{valid} sample if its training loss is large enough. We experimentally find that the percent of valid training data decreases when the network goes to the deeper classifiers. To avoid the lack of training data for the deep classifiers, we use a temperature to rescale the ensemble of the previous classifiers in order to weaken its prediction.
Third, we apply a gradient rescaling technique to the network gradients in order to bound the gradients of parameters at different layer depths. Our ablation study shows this improves the deep classifiers' performance over the baseline without gradient rescaling.

In summary, the main contributions of this paper are in three folds:
\begin{itemize}
    \item To our best knowledge, this work is the first attempt to address the \textit{train-test mismatch} problem in dynamic-depth neural networks with early exits. To this end, we formulate a dynamic neural network as an additive model, and train it in a boosting manner.
    \item To train the model effectively, we propose several techniques including multi-classifier joint optimization, prediction reweighting, and gradient rescaling techniques.
    \item The proposed BoostNet achieves the state-of-the-art performance on CIFAR100 and ImageNet datasets in both anytime and budgeted-batch prediction modes.
\end{itemize}
\section{Related Work}

\subsection{Efficient Neural Networks}
To save storage and computational resources, efficient neural networks have been widely studied in the existing works. To this end, a large number of efficient model architectures were designed in the past decade. These architectures include SqueezeNet~\cite{iandola2016squeezenet}, MobileNets~\cite{howard2017mobilenets,sandler2018mobilenetv2}, ShuffleNet~\cite{zhang2018shufflenet}, GhostNet~\cite{han2020ghostnet}, and more. Another branch of efficient methods is network pruning, which removes selected network parameters or even some entire network channels to reduce model redundancy and increase inference speed~\cite{NIPS2015_ae0eb3ee,han2015deep_compression,li2016pruning,he2017channel,luo2017thinet,he2019filter,lin2020hrank}. On the other hand, without designing or changing network architectures, the network quantization methods quantize network weights and activations from floating-point values to low-precision integers to allow fast model inference with integer arithmetic~\cite{courbariaux2016binarized,rastegari2016xnor,zhou2016dorefa,li2016ternary,zhang2018lq,liu2018bi,wang2019haq,yu2019any,liu2021post}.

\subsection{Dynamic Neural Networks}
As a special type of efficient neural networks, dynamic neural networks can adaptively change their inference complexity under different computational budgets, latency requirements, and prediction confidence requirements. Existing dynamic neural networks are designed in different aspects including sample-wise~\cite{teerapittayanon2016branchynet,wang2018skipnet,veit2018convolutional,wu2018blockdrop,yu2019any,guo2019spottune}, spatial-wise~\cite{li2017dynamic,wang2019elastic,wang2020glance,yang2020resolution,wang2022adafocus}, and temporal-wise dynamism~\cite{shen2017reasonet,yu2017learning,wu2019adaframe,meng2020ar,wang2022adafocusv3}, as categorized by \cite{han2021dynamic}.
Specially, as one type of sample-wise methods, depth-wise dynamic models with early exits adaptively exit at different layer depths given different inputs~\cite{huang2017multi,li2019improved,mcgill2017deciding,jie2019anytime,yang2020resolution}. In \cite{huang2017multi} and \cite{li2019improved}, the authors proposed multi-scale dense networks (MSDNet) to address two issues: a) interference of coarse-level and fine-level neural features, and b) feature scale inconsistency between the shallow and deep classifiers. In \cite{mcgill2017deciding} and \cite{jie2019anytime}, different inference exiting policies are proposed for better balance of accuracy and efficiency. In \cite{yang2020resolution}, resolution adaptive networks (RANet) take multi-scale images as input, based on the intuition that low-resolution features are suitable for high-efficiency inference while high-resolution features are suitable for high-accuracy inference. To train the depth-wise models, these methods jointly minimize the losses of all classifiers on every training batch without considering the dynamic inference property in testing stage. In this paper, we propose a new dynamic architecture BoostNet and several training techniques to mitigate this \textit{train-test mismatch} problem. For more literature on dynamic neural networks, we refer the readers to \cite{han2021dynamic} for a comprehensive overview on the relevant papers.

\subsection{Gradient Boosting}
As a pioneering work, AdaBoost was introduced in \cite{freund1997decision} to generate a high-accuracy model by combining multiple weak learners/models. The weak learners are optimized sequentially in a way that the subsequent learners pay more attention to the training samples misclassified by the previous classifiers.
In \cite{friedman2000additive} and \cite{friedman2001greedy}, AdaBoost was generalized to gradient boosting for different loss functions.

Later, gradient boosting methods were combined with neural networks. Following \cite{freund1997decision}, Schwenk and Bengio applied AdaBoost to neural networks for character classification~\cite{schwenk1998training}. In recent years, more investigations have been carried out to combine boosting and neural networks.
In \cite{moghimi2016boosted}, BoostCNN was introduced to combine gradient boosting and neural networks for image classification. The model achieved outperforming performance on multiple image classification tasks.
In \cite{cortes2017adanet}, the authors proposed AdaNet to learn neural network structure and parameters at the same time. The network was optimized over a generalization bound which is composed of a task loss and a regularization term.
In \cite{huang2018learning}, ResNet~\cite{he2016deep} was built sequentially in a gradient boosting way. The resulted BoostResNet required less computational and memory resources, and had exponentially decaying training errors as the layer depth increases.
More recently, GrowNet was proposed to train neural networks with gradient boosting~\cite{badirli2020gradient}, where shallow neural networks were regarded as ``weak classifiers''. In addition, after a layer is trained, a corrective finetuning step is applied to all the previous layers to improve the model performance. Different from these existing works, our training framework is designed for dynamic network structure and adaptive inference ability. To this end, we propose several optimization techniques to train the model effectively.
\section{Boosted Dynamic Neural Networks}

\subsection{Overview}
An EDNN model usually consists of multiple stacked sub-models. In inference time, given an input image, we run the model until the last layer of the first classifier to get a coarse prediction. If the prediction result is confident above a predefined threshold, the inference can be terminated at just the first classifier. Otherwise, the second classifier will be executed on the features from the first classifier. It is expected to give a more confident prediction, correcting the mistake made by the first classifier. If not, we repeat the above process until either some classifier gives a confident enough prediction or the last classifier is reached.

To train the model, existing EDNN methods train all the classifiers jointly without considering the adaptive inference property described above~\cite{huang2017multi,li2019improved,yang2020resolution}. Considering an $N$-classifier EDNN model, the overall task loss will be
\begin{equation}
L(\vx) = \sum_{n=1}^{N} w_n \ L_n(F_n(\vx), y),
\end{equation}
where $\vx$ and $y$ are an input example and its true categorical label, $F_n(\vx)$ and $w_n$ are $n$-th classifier's prediction and its contribution to the overall loss. This makes all the classifiers treat every training sample the same way. However, in inference time, the model will exit at early classifiers on easy inputs, so the deep classifiers will only be executed on difficult ones. This train-test mismatch problem will cause a data distribution gap between model training and testing stages, rendering a trained model sub-optimal.

Rethinking the inference process above, we find the intrinsic property of EDNN model is that a deeper classifier is supplementary to its previous classifiers by correcting their mistakes. This property shares a similar spirit with gradient boosting methods.

First we give a brief overview of the gradient boosting theory. Given an input $\vx$ and its output $y$, gradient boosting is to find some function $F(\vx)$ to approximate the output $y$. $F$ is in form of a sum of multiple base classifiers as
\begin{equation}
    F(\vx) = \sum_{n=1}^{N} f_n(\vx) + C,
\end{equation}
where $C$ is a constant. Gradient boosting optimizes each weak classifier sequentially:
\begin{align}
\label{equ:gradient_boosting}
F_0(\vx) &= \argmin_{\alpha} L_0(\alpha, y) \\
F_n(\vx) &= F_{n-1}(\vx) + \argmin_{f_n\in \mathcal{F}} L_n(F_{n-1}(\vx) + f_n(\vx), y), \nonumber
\end{align}
where $\alpha$ is a scalar and $\mathcal{F}$ is some function class. Directly optimizing this problem to find the optimal $f_n$ is difficult given an arbitrary loss function $L_n$. Gradient boosting methods make a first-order approximation to Eq.~\ref{equ:gradient_boosting}. As a result, we have
\begin{equation}
    f_n(\vx) = \lambda \nabla_{F_{n-1}} L_n(F_{n-1}(\vx), y),
\end{equation}
where $\lambda$ is the learning rate.

Inspired by gradient boosting, we organize the classifiers of EDNN in a similar structure. The final output of the $n$-th block is a linear combination of the outputs of $n$-th block and all its previous blocks. Fig.~\ref{fig:arch} shows the proposed network structure. We name this architecture Boosted Dynamic Neural Networks (BoostNet).

Different from the first-order approximation in gradient boosting, we directly optimize the neural network $f_n$ w.r.t. parameters $\vtheta_n$ with mini-batch gradient descent for multiple steps:
\begin{align}
    \vtheta_n^{k} &= \vtheta_n^{k-1} - \lambda \nabla_{\vtheta} L_n(\vx;\vtheta_n^{k-1}), \\
    \nabla_{\vtheta} L_n(\vx;\vtheta_n^{k-1}) &= 
    \nabla_{\vtheta} L_n(F_{n-1}(\vx) + f_n(\vx;\vtheta_n^{k-1}), y), \nonumber
\end{align}
where $F_{n-1}(\vx)$'s gradient is disabled even if it shares a subset of parameters with $f_n$ as shown in Fig.~\ref{fig:arch}. In the ablation studies, we will compare this setting with the case of differentiable $F_{n-1}(\vx)$, and find disabling $F_{n-1}(\vx)$'s gradient works better.

\begin{figure}[t]
    \centering
    \includegraphics[width=0.98\linewidth]{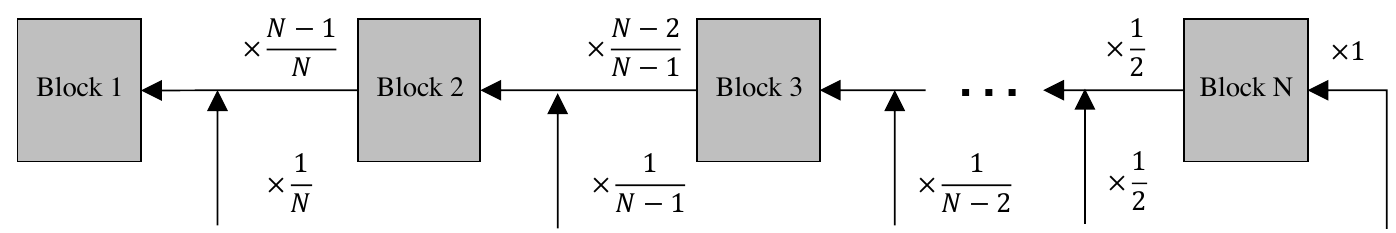}
    \caption{Gradient rescaling operation rescales gradients from different network branches by scalar multiplication. The arrows represent back-propagation directions.}
    \label{fig:gradient_reb}
\end{figure}

\begin{figure}[t]
   \centering
   \begin{subfigure}{0.49\linewidth}
     \centering
     \includegraphics[width=.99\linewidth]{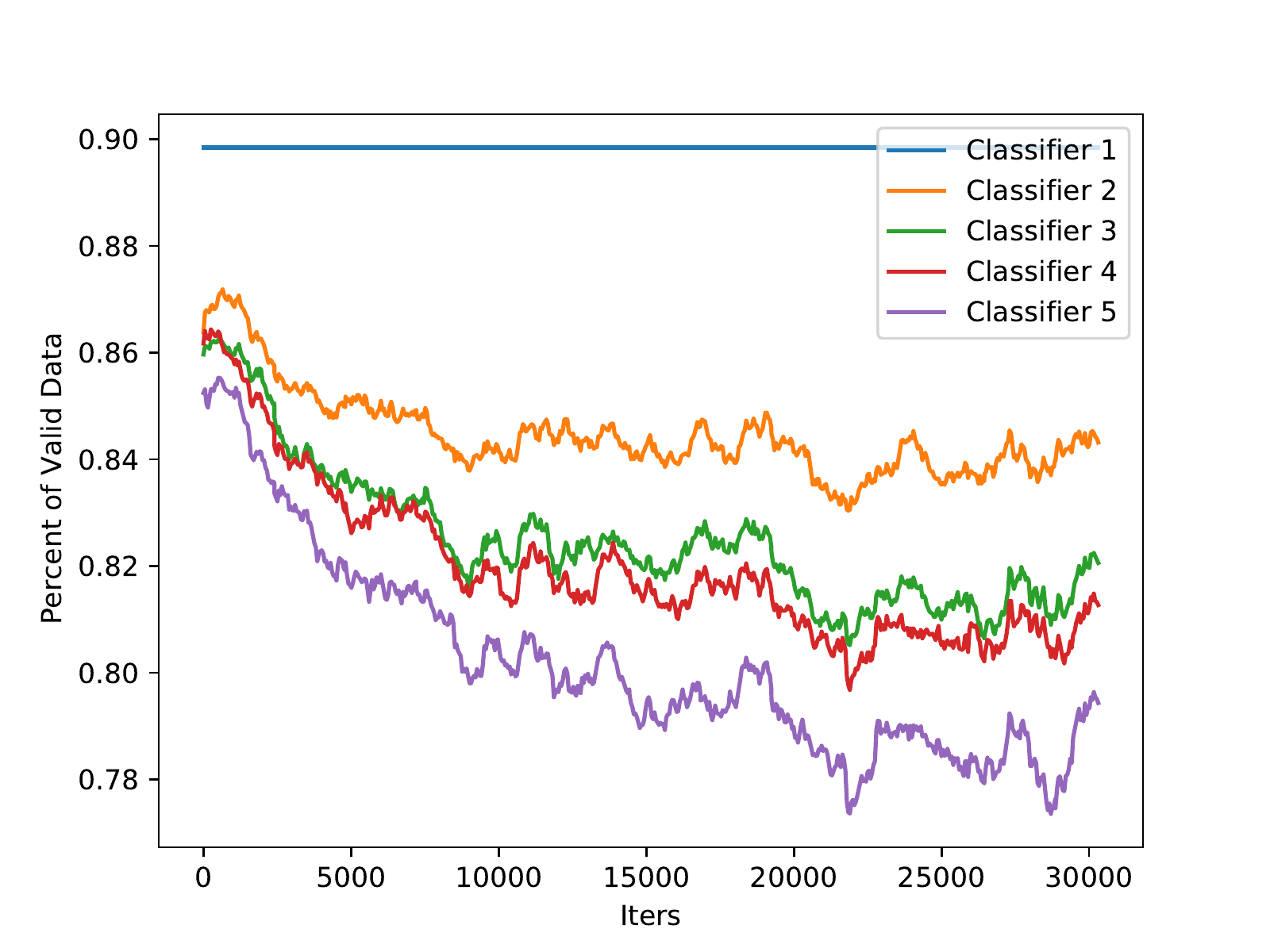}
     \caption{}
     \label{fig:valid_percent}
   \end{subfigure}
   \begin{subfigure}{0.49\linewidth}
     \centering
     \includegraphics[width=.99\linewidth]{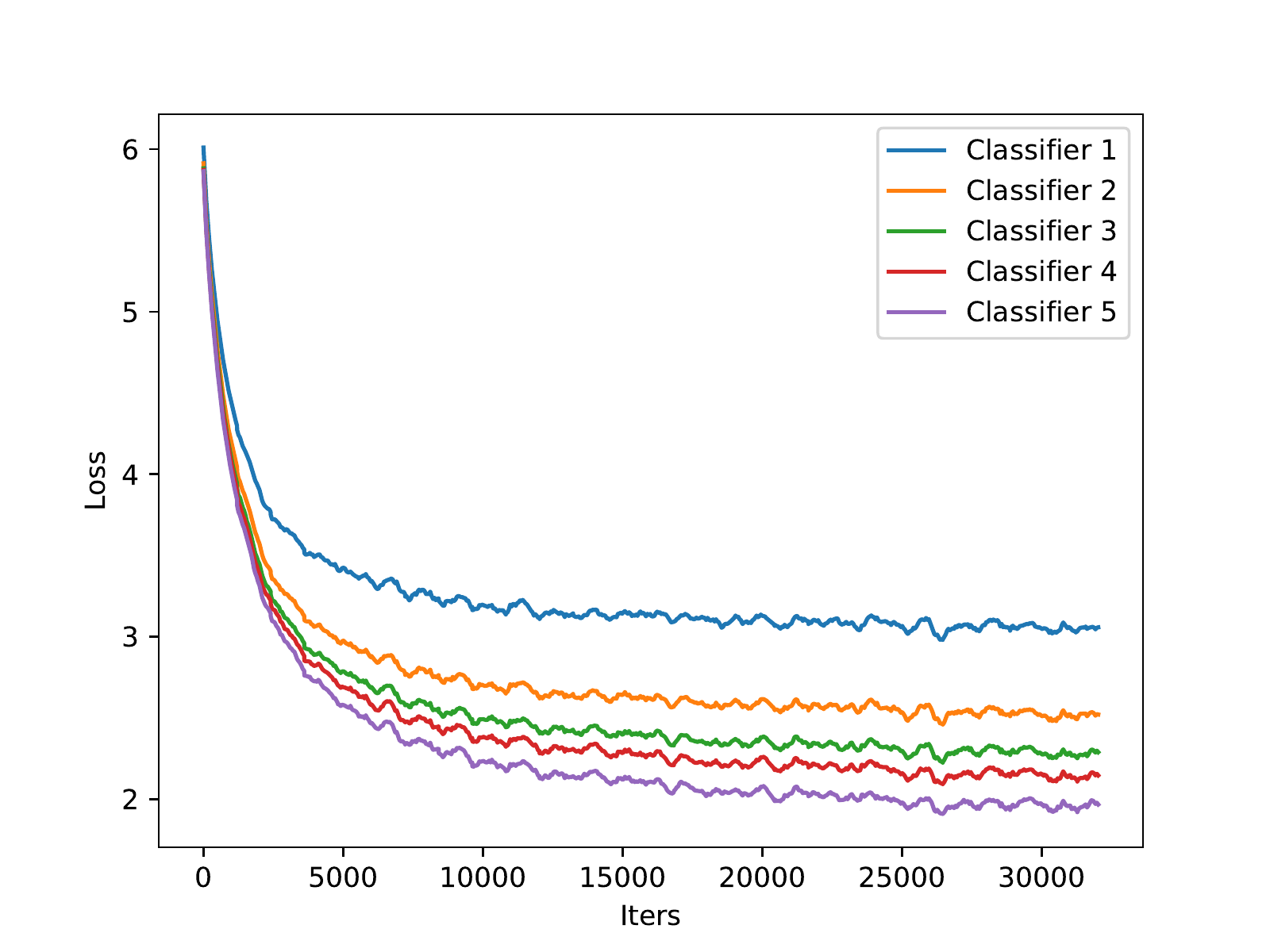}
     \caption{}
     \label{fig:loss_vs_step}
   \end{subfigure}
   \caption{Left: Percent of valid training samples for different classifiers over training iterations on ImageNet. We assume the first classifier has 90\% valid data. Right: Loss magnitudes of different classifiers over training iterations on ImageNet. We only show the first 30k training iterations.}
\end{figure}

\subsection{Mini-batch Joint Optimization}
\label{sec:joint}
Following traditional gradient boosting methods, one straight-forward optimization procedure is to train the $N$ classifiers sequentially, That is, train classifier $f_n$ until convergence, fix its classification head, and train $f_{n+1}$. However, this makes all classifiers shallower than $f_{n+1}$ deteriorate when training $f_{n+1}$ because all the classifiers share the same network backbone. On the other hand, fixing the shared parameters and only training $f_{n+1}$'s non-shared parameters will hurt its representation capacity.

Instead, we employ a mini-batch joint optimization scheme. Given a mini-batch of training data, we forward the entire BoostNet to get $N$ prediction outputs. Then we sum up the training losses over the $N$ outputs and take one back-propagation step. The overall loss is
\begin{align}
    L(\vx) &= \sum_{n=1}^{N} w_n L_n(\mathrm{stop\_grad}(F_{n-1}(\vx)) + f_n(\vx), y)
\end{align}
We experimentally find $w_n=1$ works well.

\begin{algorithm}[t]
\caption{Training pipeline for a BoostNet model.}
\label{alg:iterative_train}
\begin{algorithmic}[1]
    \Require{\ \newline
    Model $\mathcal{M}$ with parameters $\{\vtheta_n\}_{n-1}^N$ for $N$ network blocks. \newline
    Classifiers' intermediate outputs $\{\vf_n\}_{n=1}^N$. \newline
    Classifiers' final outputs $\{\mF_n\}_{n=1}^N$. \newline
    Prediction temperatures $\{t_n\}|_{n=1}^{N}$. \newline
    Training dataset $\mX = \{(\vx, y)\}$.}
    \Ensure{\ \newline
    Model $\mathcal{M}$}
    \State $\mF_0 \gets 0$
    \For{$t \gets 1$ to $T$}  \Comment{Iterate over training batches}
      \State $l \gets 0$
      \For{$n \gets 1$ to $N$}  \Comment{Iterate over classifiers}
        \State $\mF^{\prime} \gets$ stop\_gradient($t_n \mF_{n-1}(\vx_t)$)
        \State $l_n \gets$ loss\_function($\mF^{\prime} + \vf_n(\vx_t)$, $\vy_t$)
        \State $l \gets l + l_n$
      \EndFor
        \State [$\vg_1, ..., \vg_n$] $\gets$ compute\_gradients($l$, [$\vtheta_1$, ..., $\vtheta_n$])
        \State apply\_gradients([$\vtheta_1$, ..., $\vtheta_n$], [$\vg_1, ..., \vg_n$])
    \EndFor
    \State return $\mathcal{M}$
\end{algorithmic}
\end{algorithm}

\subsection{Gradient Rescaling}
Similar to \cite{li2019improved}, we use a gradient rescaling operation to rescale the gradients from different network branches. As shown in Fig.~\ref{fig:gradient_reb}, we multiply a scalar in each branch to rescale the gradients that pass through it. With this operation, the gradient of the overall loss $L$ w.r.t. block $b_n$ is
\begin{align}
    \frac{\partial L}{\partial b_n} &= \frac{1}{N-n+1} \sum_{i=n}^{N} \frac{\partial L_i}{\partial b_n},
\end{align}
which is bounded even if $N$ is a large number, given the weak assumption that $\partial L_i / \partial b_n$ is bounded, which is usually true. The effectiveness of gradient rescaling will be demonstrated in the ablation studies.

\begin{table*}[th]
    \centering
    \scriptsize
    \begin{tabular}{l|C{0.4cm}C{0.4cm}C{0.4cm}C{0.4cm}C{0.4cm}C{0.4cm}C{0.4cm}C{0.4cm}C{0.4cm}C{0.5cm}|C{0.4cm}C{0.4cm}C{0.4cm}C{0.4cm}C{0.4cm}C{0.4cm}C{0.4cm}C{0.4cm}}
    \hline
    & \multicolumn{10}{c|}{MSDNet} & \multicolumn{8}{c}{RANet}\\
    \hline
         Classifier index & 1 & 2 & 3 & 4 & 5 & 6 & 7 & 8 & 9 & 10 & 1 & 2 & 3 & 4 & 5 & 6 & 7 & 8 \\ 
     \hline
         Mul-Add ($\times 10^{6}$) & 15.14& 25.51& 37.79& 52.06& 62.18& 74.07& 87.73& 100.0& 110.1& 121.7 & 15.83& 31.31& 44.69& 50.40& 60.71& 63.72& 90.26& 95.00 \\
         Baseline & 64.60 & 67.94& 69.36& 71.53& 72.76& 74.03& 74.65& 75.37& 75.53& 75.98 & 66.71& 69.11& 71.22& 71.99& 72.86& 72.93& 73.88& 74.69 \\
         Ours & 65.32& 70.01& 72.34& 74.15& 75.24& 75.85& 75.94& 76.51& 76.53& 76.30 & 67.53& 70.83& 73.17& 73.39& 74.20& 74.05& 74.75& 75.08 \\
    \hline
    \end{tabular}
    \caption{Comparison on CIFAR100 in anytime prediction mode. In this mode, the model runs at a fixed computational budget for all inputs. We compare our method with MSDNet and RANet using the corresponding backbones.}
    \label{tab:any_mode_cifar100}
\end{table*}

\begin{table*}[th]
    \centering
    \scriptsize
    \begin{tabular}{l|C{0.4cm}C{0.4cm}C{0.4cm}C{0.4cm}C{0.4cm}C{0.4cm}C{0.4cm}C{0.5cm}|C{0.4cm}C{0.4cm}C{0.4cm}C{0.4cm}C{0.4cm}}
    \hline
    & \multicolumn{8}{c|}{MSDNet} & \multicolumn{5}{c}{RANet}\\
    \hline   
    Classifier index & 1 & 2 & 3 & 4 & 5 & 6 & 7 & 8 & 1 & 2 & 3 & 4 & 5\\ 
    \hline
         Mul-Add ($\times 10^{6}$) & 435.0& 888.7& 1153& 1352& 2280& 2588& 3236& 3333 & 615.7& 1436& 2283& 2967& 3254\\
         Baseline & 63.29& 67.59& 69.90& 70.84& 74.25& 74.73& 74.29& 74.46 & 63.69& 71.12& 73.60& 74.46& 75.24\\
         Ours & 64.19& 68.33& 71.07& 71.53& 74.54& 74.90& 75.09& 74.66 & 64.78& 71.63& 74.48& 75.35& 76.05\\
    \hline
    \end{tabular}
    \caption{Comparison on ImageNet in anytime prediction mode. MSDNet and RANet are compared.}
    \label{tab:any_mode_imagenet}
\end{table*}

\subsection{Prediction Reweighting with Temperature}
\label{sec:temperature}

When training in the above procedure, we observed a problem that the deeper classifiers do not show much performance improvement over the shallow classifiers, although the model size increases by a large margin.

We found this is caused by insufficient training data for deeper classifiers. In Fig.~\ref{fig:valid_percent}, we show the number of valid training samples for each classifier. A training sample is valid if its training loss is larger than a certain threshold $v$. Here, we set $v$ to be the 10-th percentile of all the training losses of the first classifier in each batch of training data, i.e., we assume the first classifier always has $90\%$ valid training samples. From the figure, we find the shallower classifiers have more valid training data while the deeper classifiers have relatively less. In Fig.~\ref{fig:loss_vs_step}, we show the batch-wise average training losses of different classifiers. Consistent with Fig.~\ref{fig:valid_percent}, the deeper classifiers have smaller training losses. From another perspective, this can be explained by the fact that the deeper classifiers have a larger model size and thus more representation power, but this also indicates that the deeper classifiers need more challenging training samples instead of learning the ``easy'' residuals of the shallower classifiers.

To address this problem, we multiply the output of $(n-1)$-th classifier by a temperature $t_n$ before serving as an ensemble member of $F_n$ as
\begin{equation}
    F_n = t_n F_{n-1} + f_n,
\end{equation}
where $F_{n-1}$ and $F_n$ are pre-softmax prediction logits. We set all the temperatures $\{t_n\}|_{n=1}^{N}$ to $0.5$ in our experiments unless specified in the ablation studies. We name this technique \textit{prediction reweighting}. With this, the ensemble output of the shallow classifiers $F_{n-1}$ is equivalent to being weakened before being used in the ensemble output $F_n$.

With the proposed training techniques, the overall training pipeline is summarized in Algorithm~\ref{alg:iterative_train}.
\section{Experiments}
\subsection{Settings}
To evaluate the proposed method, we apply it to two different network backbones, MSDNet~\cite{huang2017multi} and RANet~\cite{yang2020resolution}, both of which are dynamic models designed for adaptive inference.
It is noted that our method is model-agnostic, and can also be applied to other network backbones.
We do experiments on two datasets, CIFAR100~\cite{Krizhevsky09learningmultiple} and ImageNet~\cite{deng2009imagenet} , in two prediction modes, anytime prediction mode and budgeted-batch prediction mode following \cite{huang2017multi,yang2020resolution}. In anytime prediction mode, the model runs at a fixed computational budget for all inputs. In budgeted-batch prediction mode, the model runs at an average budget, which means it can have different inference costs for different inputs as long as the overall average cost is below a threshold. To calculate the confidence threshold determining whether the model exits or not, we randomly reserve 5000 samples in CIFAR100 and 50000 samples in ImageNet as a validation set. We utilize a similar model combination strategy to \cite{huang2017multi,yang2020resolution}. That is, we train a single dynamic model for anytime mode on both datasets. For budgeted-batch mode, we train multiple models with different model sizes for each dataset.

For MSDNet~\cite{huang2017multi}, we use the same model configurations as the original model except the CIFAR100-Batch models. In \cite{huang2017multi}, the classifiers are located at depths $\{2\times i+4|i=1,...,k\}$, where $k$ is $6$, $8$, or $10$ for the three different models. One problem with this setting is that the three models have the same representation capacity for the shallow classifiers, which causes redundant overlap of computational complexity. Instead, we set the locations as $\{2\times i+k|i=1,...,6\}$, where $k$ is $4$, $8$, or $12$ for the three models. Thus, each model has $6$ classifiers at different depths. For RANet~\cite{yang2020resolution}, we follow the same settings as the original paper. For prediction reweighting, we set all temperatures $t_n=0.5,\ n=1,...,N$.

\begin{figure*}[t]
   \centering
   \begin{subfigure}[t]{0.24\textwidth}
        \centering
        \includegraphics[width=0.99\linewidth]{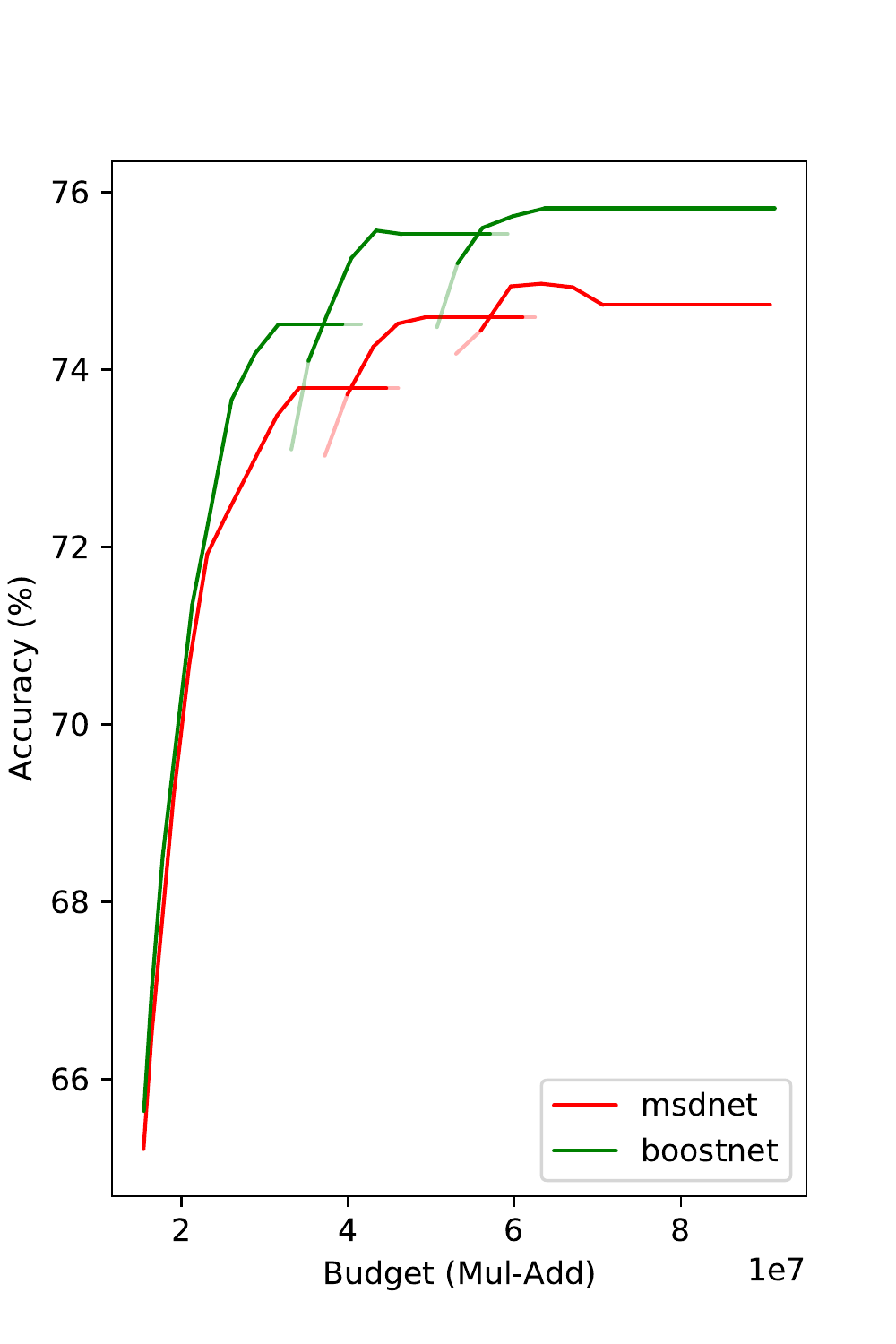}
        \caption{CIFAR100, MSDNet}
        \label{fig:batch_cifar100}
    \end{subfigure}
   \begin{subfigure}[t]{0.24\textwidth}
        \centering
        \includegraphics[width=0.99\linewidth]{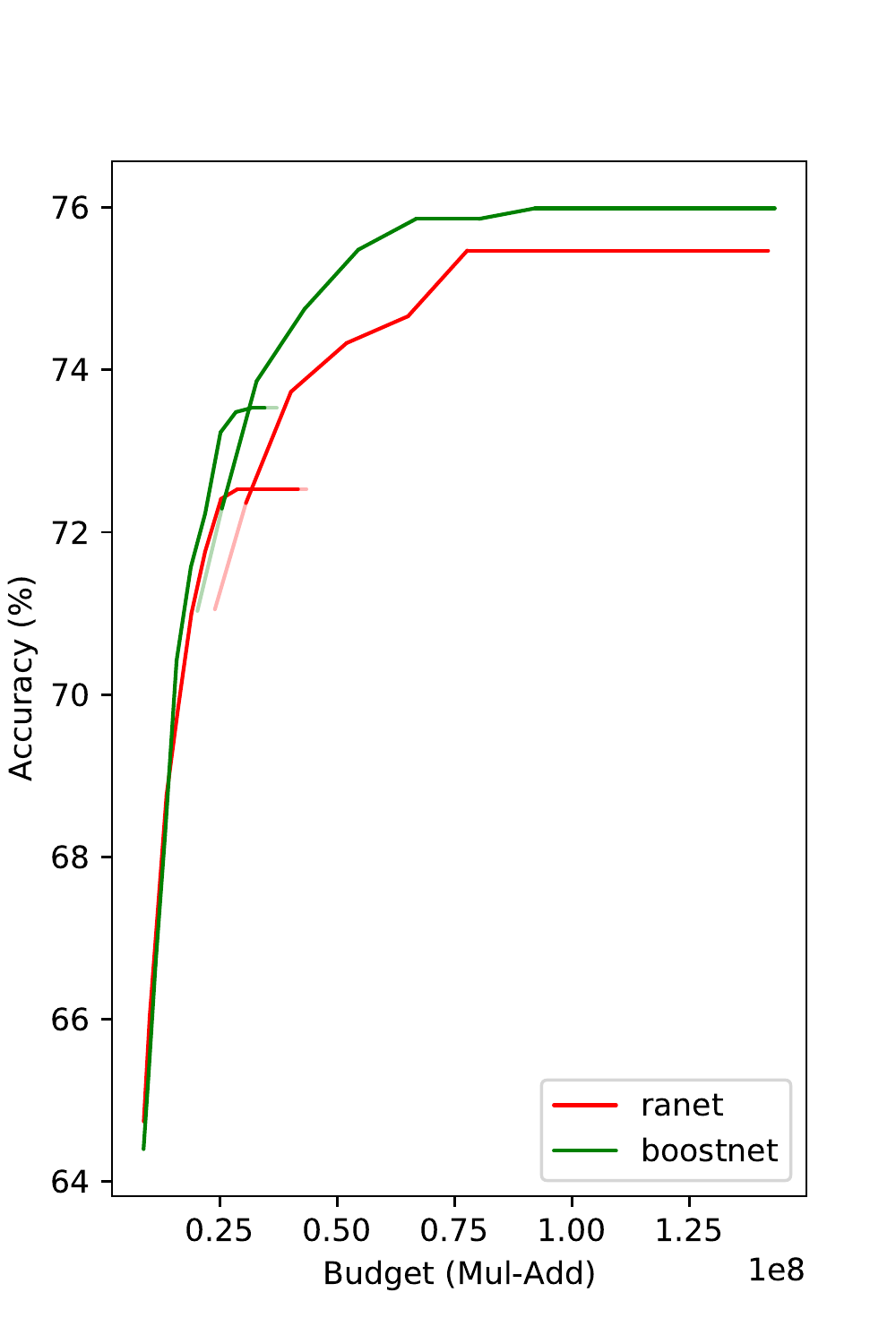}
        \caption{CIFAR100, RANet}
        \label{fig:batch_cifar100_ra}
    \end{subfigure}
   \begin{subfigure}[t]{0.24\textwidth}
        \centering
        \includegraphics[width=0.99\linewidth]{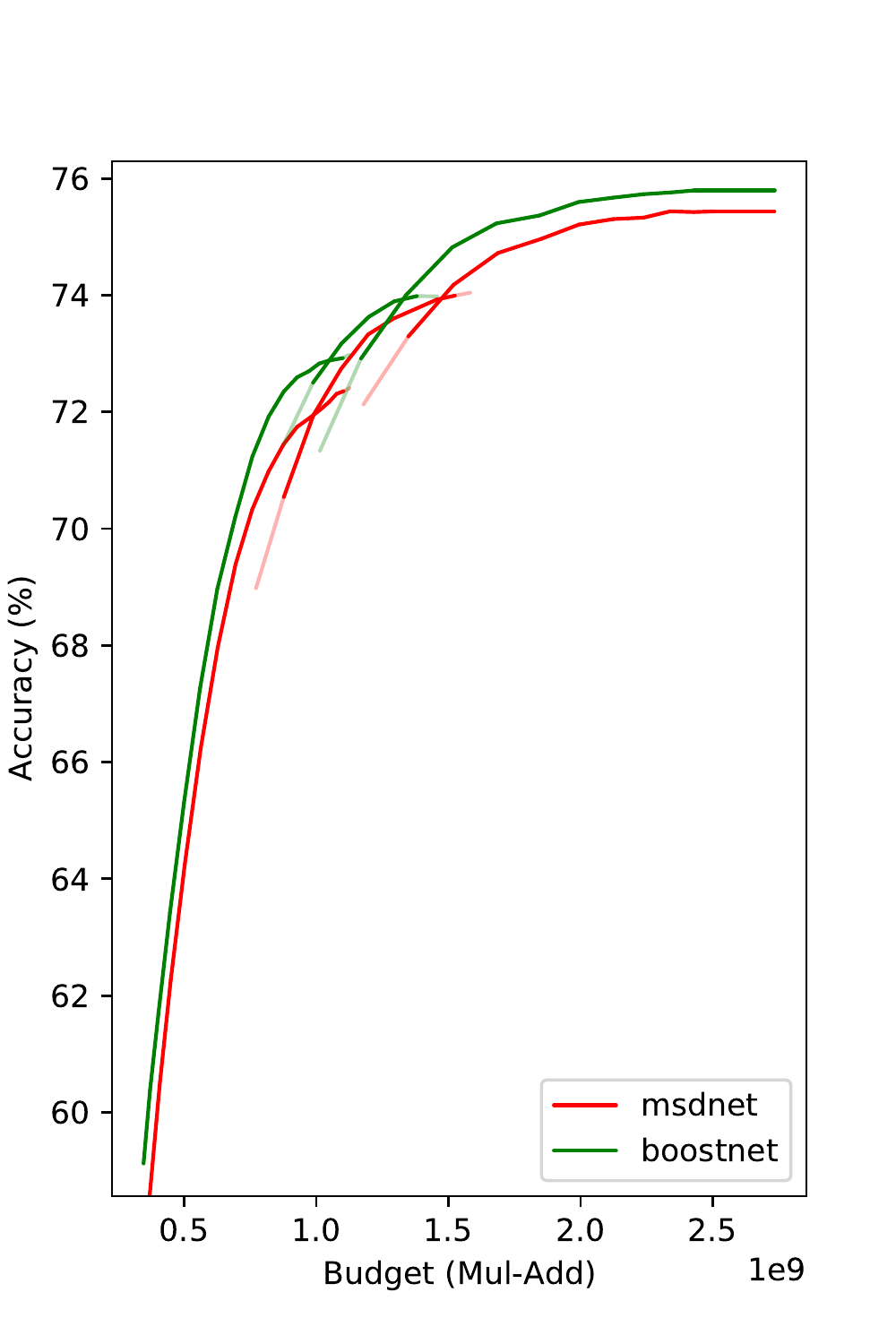}
        \caption{ImageNet, MSDNet}
    \end{subfigure}
   \begin{subfigure}[t]{0.24\textwidth}
        \centering
        \includegraphics[width=0.99\linewidth]{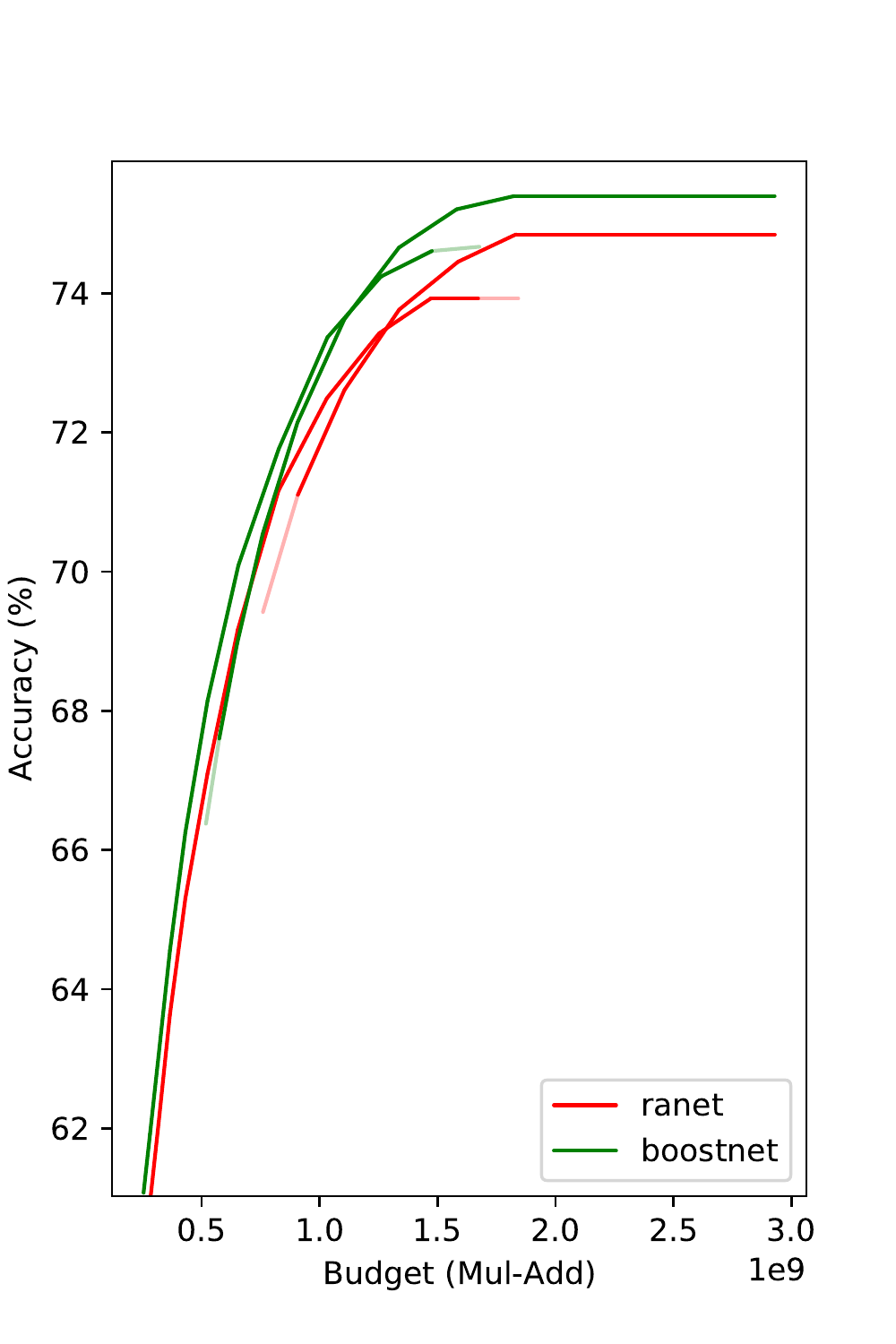}
        \caption{ImageNet, RANet}
    \end{subfigure}
    \caption{Comparison in budgeted-batch prediction mode. In this mode, the model runs at an average computational budget, which means it can have different inference costs for different inputs as long as the overall average cost is below a threshold. We compare our method with MSDNet and RANet using the corresponding backbones.}
    \label{fig:batch_mode}
\end{figure*}

For CIFAR100, the models are trained for $300$ epochs with SGD optimizer with momentum $0.9$ and initial learning rate $0.1$. The learning rate is decayed at epochs $150$ and $225$. We use single GPU with training batch size $64$. 
For ImageNet, the models are trained for $90$ epochs with SGD optimizer with momentum $0.9$ and initial learning rate $0.1$. The learning rate is decayed at epochs $30$ and $60$. We use $4$ GPUs with training batch size $64$ on each GPU. Our training framework is implemented in PyTorch~\cite{NEURIPS2019_9015}.

\subsection{Anytime Prediction Mode}
In this section, we compare our method with MSDNet and RANet in the anytime prediction mode. We also apply gradient rescaling to the baseline models. The results on CIFAR100 and ImageNet are plotted in
Tab.~\ref{tab:any_mode_cifar100} and \ref{tab:any_mode_imagenet}.
On both datasets, our model achieves better results under the same computational budget measured in {Multi-Add}. The improvement is even more significant under the low computational budgets on CIFAR100.

\subsection{Budgeted-batch Prediction Mode}
In budgeted-batch prediction mode, we set a computational complexity upper bound $\tau$ such that the average computational cost of the input images is below this value. We assume a proportion of $\frac{p}{1}$ samples exit at every classifier. Therefore, the average computational cost will be 
\begin{equation}
C_{avg}=\sum_{n=1}^{N}(1-p)^{n-1}C_n,
\end{equation}
where $N$ is the number of classifiers and $C_n$ is the computational complexity of layers between block $n-1$ and $n$. Given a budget threshold, we can utilize the Newton–Raphson method to find a $p$ satisfying the equation. Finally, we run the model on the hold-out evaluation set to determine the confidence threshold for each classifier given exiting probability $p$. We define the confidence of a classifier as its maximum softmax probability over all the categories.

We compare our method with MSDNet and RANet in Fig.~\ref{fig:batch_mode}.\footnote{Due to various reasons, \textit{e.g.}, different validation sets, RANet's CIFAR100 result here may be a bit different from that in the paper.} Under the same budget, our method gets superior results. On CIFAR100, the improvement over MSDNet is even more than $1\%$.
We observe that the improvement decreases as the budget increases. We speculate the reason is when the model goes deeper, the effective amount of training data decreases. The proposed prediction reweighting technique mitigates this problem, but it is not entirely eliminated.

Note that the curve tails are flat in Fig.~\ref{fig:batch_cifar100} and Fig.~\ref{fig:batch_cifar100_ra}, \textit{i.e.}, a deeper classifier does not improve the performance over a shallower one.
One possible reason of such a phenomenon of diminishing returns is overfitting.
To avoid performance drop when computational budget increases, we do not increase confidence threshold if this change does not bring performance improvement. This process is conducted on the hold-out evaluation dataset. A similar strategy is also employed in \cite{yang2020resolution}.

\subsection{Inference Visualization}
\begin{figure*}[t]
    \centering
    \begin{subfigure}[t]{0.244\textwidth}
        \centering\includegraphics[width=0.99\linewidth]{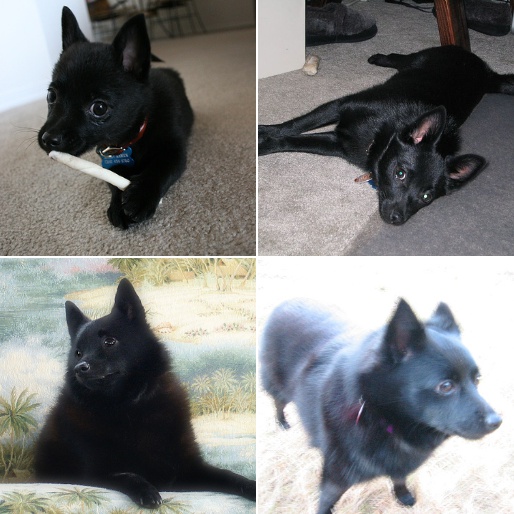}
        \caption{{\scriptsize Schipperkes at cls. 1}}
        \label{fig:visual_1}
    \end{subfigure}
    \begin{subfigure}[t]{0.244\textwidth}
        \centering\includegraphics[width=0.99\linewidth]{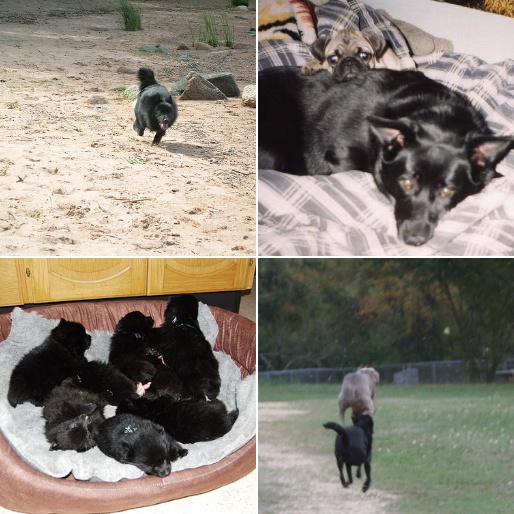}
        \caption{{\scriptsize Schipperkes at cls. 5}}
        \label{fig:visual_2}
    \end{subfigure}
    \begin{subfigure}[t]{0.244\textwidth}
        \centering\includegraphics[width=0.99\linewidth]{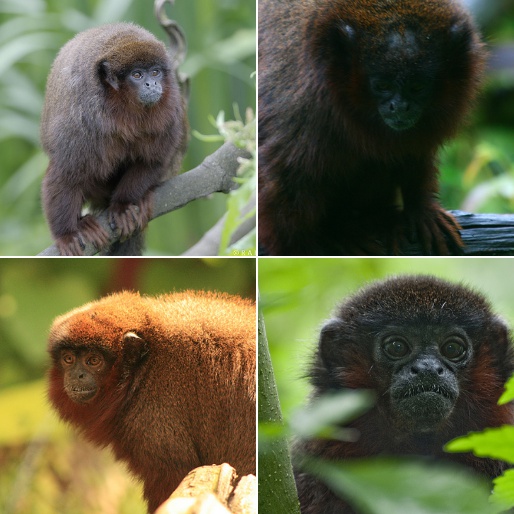}
        \caption{Titis at cls. 1}
        \label{fig:visual_3}
    \end{subfigure}
    \begin{subfigure}[t]{0.244\textwidth}
        \centering\includegraphics[width=0.99\linewidth]{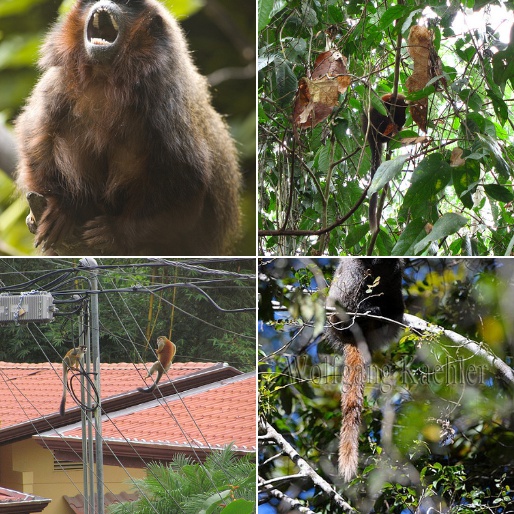}
        \caption{Titis at cls. 5}
        \label{fig:visual_4}
    \end{subfigure}
    \caption{Visual examples from ImageNet exiting at classifiers $1$ and $5$. Left: schipperke (id 223). Right: titi (id 380).}
    \label{fig:visual}
\end{figure*}

We show visual examples that exit at various depths of the dynamic model. Fig.~\ref{fig:visual_1} and \ref{fig:visual_2} are schipperke images that exit at classifier $1$ and $5$ respectively. Fig.~\ref{fig:visual_3} and \ref{fig:visual_4} are titi examples exiting at classifier $1$ and $5$ respectively. We find that image instances exiting at classifier $1$ are usually ``easier'' than those exiting at classifier $5$. Actually, this phenomenon can be observed in all the categories in ImageNet. Generally, images with occlusion, small object size, unusual view angle, and complex background are difficult to recognize and require a deeper classifier.

\subsection{Ablation Studies}
\label{sec:ablation}
\begin{figure*}[t]
    \centering
    \begin{subfigure}[t]{0.3\textwidth}
        \centering\includegraphics[width=0.99\linewidth]{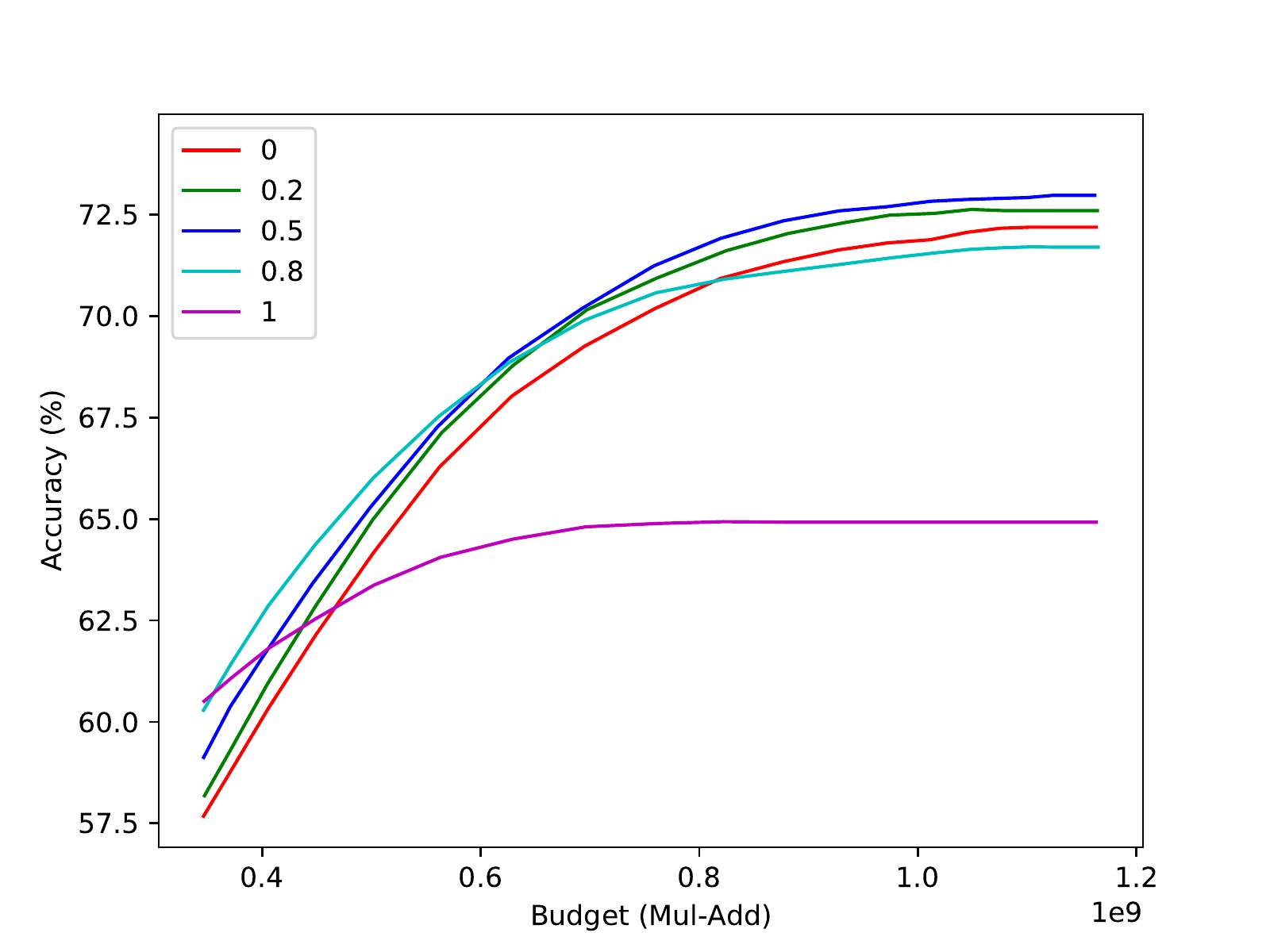}
        \caption{}
        \label{fig:ablation_t}
    \end{subfigure}
    \begin{subfigure}[t]{0.3\textwidth}
        \centering\includegraphics[width=0.99\linewidth]{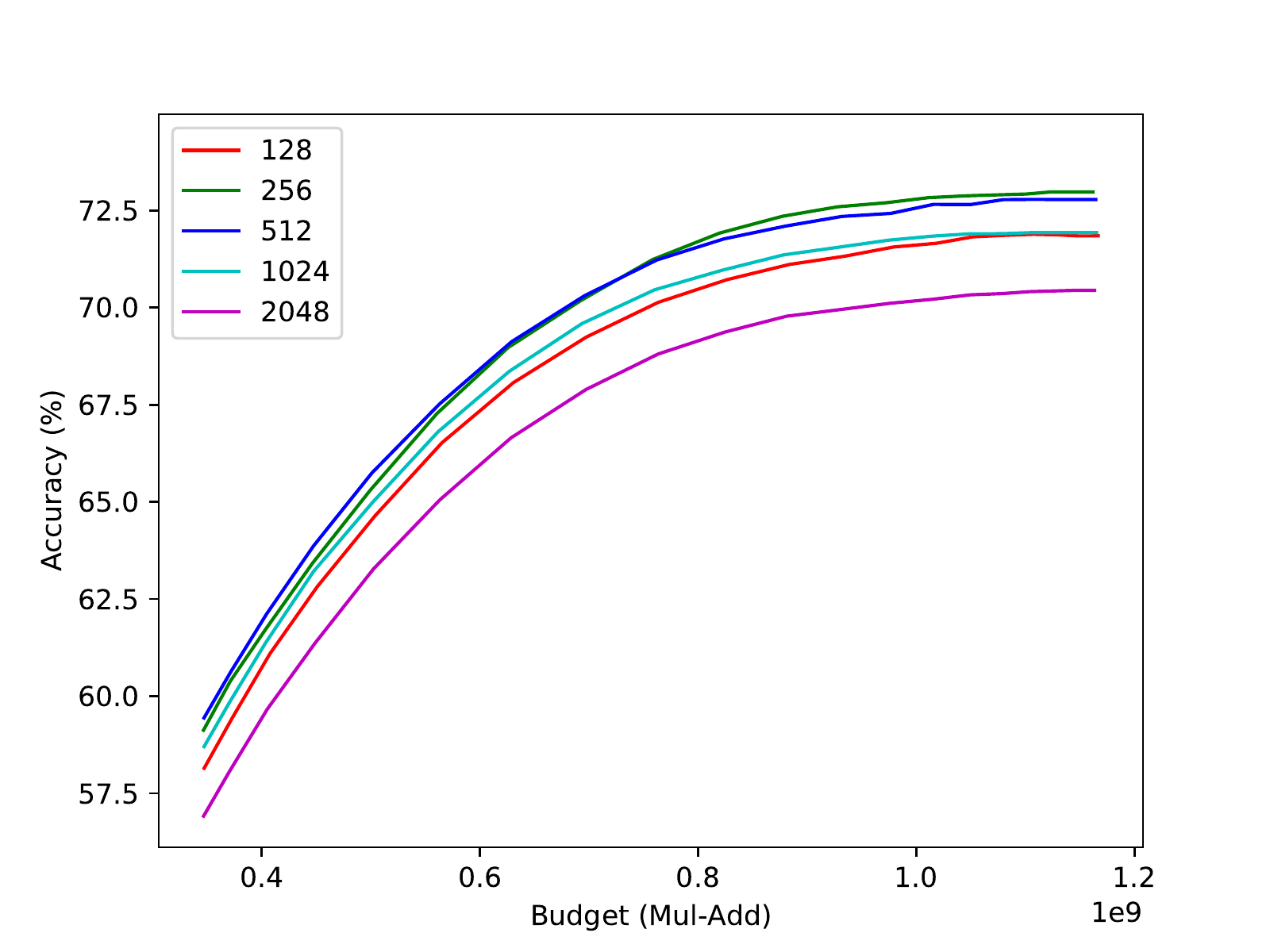}
        \caption{}
        \label{fig:ablation_bs}
    \end{subfigure}
    \begin{subfigure}[t]{0.3\textwidth}
        \centering\includegraphics[width=0.99\linewidth]{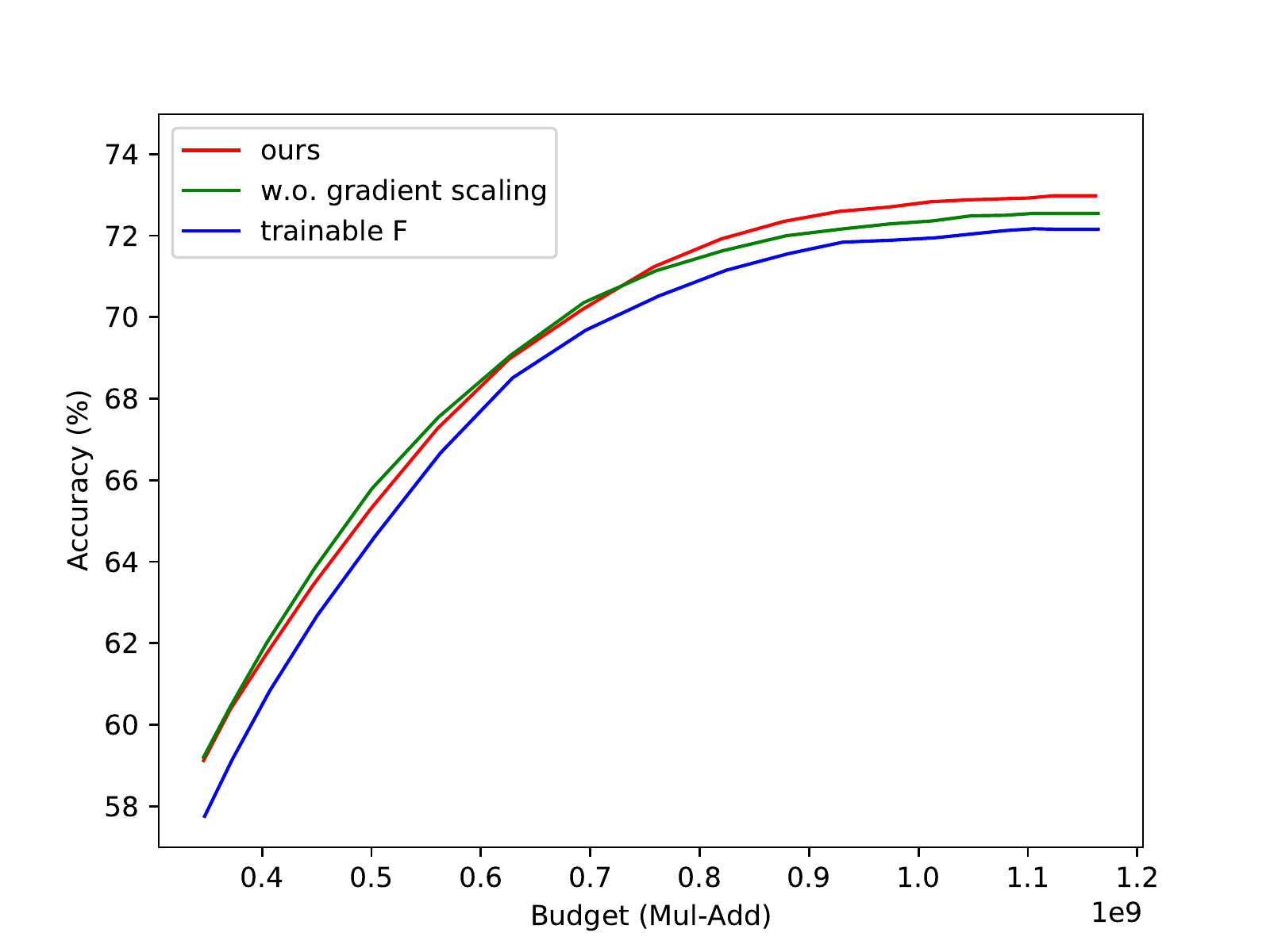}
        \caption{}
        \label{fig:ablation_other}
    \end{subfigure}
    \caption{Ablation studies on different training settings. Left: Ablation study on different temperatures. We use $t=0.5$ in all experiments by default. Middle: Ablation study on different batch sizes. We use $batch\_size=256$ in all experiments by default. Right: Ablation studies on gradient rebalancing and fixing early model ensemble.}
    \label{fig:ablation}
\end{figure*}

In this section, we do ablation experiments using MSDNet backbone to understand the training techniques.

\textbf{Prediction reweighting}. In the previous section, we introduced prediction reweighting to make sure more valid training data for deeper classifiers. In this part, we study how prediction reweighting temperature $t$ affects the performance. The results are plotted in Fig.~\ref{fig:ablation_t}. We find that the deeper classifiers cannot learn well without enough supervision when $t$ is $1$, although the shallower classifiers achieve better performance than the other $t$ values. On the other hand, when $t$ is $0$, it basically becomes an MSDNet, which also performs inferior to the case of $t=0.5$. As a trade-off, we set $t=0.5$ in all our experiments.

\textbf{Batch size}. As introduced earlier, we formulated the early-exiting dynamic-depth neural network as an additive model in a way that the prediction of a deep classifier $F_n$ depends on all its preceding classifiers $\{F_i\}_{i=1}^{n-1}$. At training step $k$, $F_n^k$ is computed using $\{F_i^{k-1}\}_{i=1}^{n-1}$ before the new $\{F_i^{k}\}_{i=1}^{n-1}$ are computed. Intuitively thinking, if these shallower classifiers fluctuate too much during training, the deeper classifier cannot be well learned. Therefore, a relatively large batch size is desired for better training stability. Meanwhile, it cannot be too large to hurt the model's generalization ability~\cite{keskar2016large}. Therefore, batch size plays an important role in dynamic network training. In Fig.~\ref{fig:ablation_bs}, we show how batch size affects the classification performance. Both too large batch sizes and too small batch sizes get worse performance. Therefore, we set the batch size to be $64$ for CIFAR100 models and $256$ for ImageNet models as a trade-off between stability and generalization.

\textbf{Gradient rescaling}. In Fig.~\ref{fig:ablation_other}, we compare the model with gradient rescaling (red curve) and that without gradient rescaling (green curve). When the budget is over $0.8\times10^{9}$, the model with gradient rescaling gets better performance than that without it. This shows gradient rescaling can help improve the performance of the deeper classifiers.

\textbf{Trainable $F_{n-1}$}. As shown in Algorithm~\ref{alg:iterative_train}, we do not back-propagate gradients of $l_n$ along $F_{n-1}$ branch. We do this to avoid entangling deeper losses in shallower classifiers. For comparison purpose, we train a model without cutting out the gradients of $l_n$ w.r.t. $F_{n-1}$. The results are shown as the blue curve in Fig.~\ref{fig:ablation_other}. Entangling deeper losses in shallower classifiers hurts classification performance.
\section{Discussion}
To analyze the training process of the deep classifiers in the proposed BoostNet, we introduced the concept of effective training data in dynamic network training. It measures the proportion of training samples that have large enough losses for effectively training a classifier. To make a trade-off between solving train-test mismatch problem and increasing the proportion of effective training data, we introduced a prediction reweighting temperature to weaken the prediction of the early classifiers. One future research topic is to design a better metric to evaluate the amount of effective training data, and a better prediction reweighting algorithm. For instance, use a dynamic temperature that adaptively adjusts the effective training data amount for different classifiers on each training batch.

Dynamic-depth neural networks can be regarded as a combination of multiple sub-models, each of which concentrates on a subset of training data. Existing methods train all the sub-models on all the training data. Our method trains the first sub-model on all the data, and deeper sub-models on the hard samples. Another possibility is to train different sub-models on different subsets, avoiding training the shallower classifiers on the hard samples and the deeper classifiers on the easy samples. In this case, a model selection module is needed to determine which sub-model should be used for an input. A similar idea is explored in \cite{zhang2021collaboration}, where an expert selection module is proposed to do expert (sub-model) selection. However, their model is not designed for adaptive inference, and the low-level feature extraction module is not shared among the experts for better inference efficiency.

Our method has some limitations. First, BoostNet is mainly designed for image classification tasks. For other tasks like object detection and semantic segmentation, it would be difficult to determine if an input is easy or difficult at the image level since we make pixel-level or region-level predictions. Applying our method to other vision tasks is still under exploration. Second, there exists an efficiency gap between theoretical analysis and real-world application on different hardware. Traditional neural networks can be executed in parallel via batch inference, but the adaptive inference models like the proposed BoostNet need a special design to allow better batch inference since the model has different inference paths on different inputs.

\section{Conclusion}
In this paper, we introduced a Boosted Dynamic Neural Network architecture. Different from existing dynamic-depth neural networks, we formulate our dynamic network as an additive model similar to gradient boosting in order to solve the \textit{train-test mismatch} problem. To optimize the model effectively, we jointly update all the classifiers at the same time, which avoids the classifiers falling into sub-optimal solutions. In addition, we propose prediction reweighting with temperature to increase the amount of effective training data for deep classifiers. To bound the gradients from all classifiers, we leverage gradient rescaling to rescale the gradients from each branch. Our method achieves superior performance on CIFAR100 and ImageNet datasets in both anytime and budgeted-batch prediction modes.

\bibliography{7_references}

\end{document}